# Regret-based Reward Elicitation for Markov Decision Processes

**Kevin Regan**
Department of Computer Science
University of Toronto
Toronto, ON, CANADA
kmregan@cs.toronto.edu

**Craig Boutilier**
Department of Computer Science
University of Toronto
Toronto, ON, CANADA
cebly@cs.toronto.edu

## Abstract

The specification of a Markov decision process (MDP) can be difficult. Reward function specification is especially problematic; in practice, it is often cognitively complex and time-consuming for users to precisely specify rewards. This work casts the problem of specifying rewards as one of preference elicitation and aims to minimize the degree of precision with which a reward function must be specified while still allowing optimal or near-optimal policies to be produced. We first discuss how *robust policies* can be computed for MDPs given only partial reward information using the minimax regret criterion. We then demonstrate how regret can be reduced by efficiently eliciting reward information using bound queries, using regret-reduction as a means for choosing suitable queries. Empirical results demonstrate that regret-based reward elicitation offers an effective way to produce near-optimal policies without resorting to the precise specification of the entire reward function.

## 1 Introduction

Markov decision processes (MDPs) have proven to be an extremely useful formalism for decision making in stochastic environments. However, the specification of an MDP by a user or domain expert can be difficult, e.g., cognitively demanding, computationally costly, or time consuming. For this reason, much work has been devoted to learning the dynamics of stochastic systems from transition data, both in offline [11] and online (i.e., reinforcement learning) settings [19]. While model dynamics are often relatively stable in many application domains, MDP reward functions are much more variable, reflecting the preferences and goals of specific users in that domain. This makes reward function specification more difficult: they can't generally be specified *a priori*, but must be elicited or otherwise assessed for individual users. Even online RL methods require the specification of a user's reward function in some form: unlike state transitions, it is impossible to directly observe a reward function except in very specific settings with simple, objectively definable, observable performance criteria. The "observability" of reward is a convenient fiction often assumed in the RL literature.

Reward specification is difficult for three reasons. First, it requires the translation of user preferences—which states and actions are "good" and "bad"—into precise numerical rewards. As has been well-recognized in decision analysis, people find it extremely difficult to quantify their strength of preferences precisely using utility functions (and, by extension, reward functions) [10]. Second, the requirement to assess rewards and costs for *all* states and actions imposes an additional burden (one that can be somewhat alleviated by the use of multiattribute models in factored MDPs [5]). Finally, the elicitation problem in MDPs is further exacerbated by the potential conflation of immediate reward (i.e., $r(s, a)$) with long-term value (either $Q(s, a)$ or $V(s)$): states can be viewed as good or bad based on their ability to make other good states reachable.

In this paper, we tackle the problem of reward elicitation in MDPs by treating it as a preference elicitation problem. Recent research in preference elicitation for non-sequential decision problems exploits the fact that optimal or near-optimal decisions can often be made with relatively imprecise specification of a utility function [6, 8]. Interactive elicitation and optimization techniques take advantage of feasibility restrictions on actions or outcomes to focus their elicitation efforts on only the most relevant aspects of a utility function. We adopt a similar perspective in the MDP setting, demonstrating that optimal and near-optimal policies can be often found with limited reward information. For instance, reward bounds in conjunction with MDP dynamics can render certain regions of state space provably dominated by others (w.r.t. value).

We make two main contributions that allow effective elicitation of reward functions. First, we develop a novel robust optimization technique for solving MDPs with imprecisely specified rewards. Specifically, we adopt the *minimax regret* decision criterion [6, 18] and develop a formulation for MDPs: intuitively, this determines a policy that has minimum regret, or loss w.r.t. the optimal policy, over all possible reward function realizations consistent with the cur-rent partial reward specification. Unlike other work on robust optimization for imprecisely specified MDPs, which focuses on the maximin decision criterion [1, 13, 14, 16], minimax regret determines superior policies in the presence of reward function uncertainty. We describe an exact computational technique for minimax regret and suggest several approximations. Second, we develop a simple elicitation procedure that exploits the information provided by the minimax-regret solution to guide the querying process. In this work, we focus on simple schemes that refine the upper and lower bounds of specific reward values. We show that good or optimal policies can be determined with very imprecise reward functions when elicitation effort is focused in this way. Our work thus tackles the problem of reward function precision directly. While we do not address the issue of reward-value conflation in this model, we will discuss it further below.

## 2 Notation and Problem Formulation

We begin by reviewing MDPs and defining the minimax regret criterion for MDPs with imprecise rewards.

### 2.1 Markov Decision Processes

Let $\langle S, A, \{P_{sa}\}, \gamma, \alpha, r\rangle$ be an infinite horizon MDP with: finite state set $S$ of size $n$; finite action set $A$ of size $k$; transition distributions $P_{sa}(\cdot)$, with $P_{sa}(t)$ denoting the probability of reaching state $t$ when action $a$ is taken at $s$; reward function $r(s,a)$; discount factor $\gamma < 1$; and initial state distribution $\alpha(\cdot)$. Let $\mathbf{r}$ be the $n \times k$-vector with entries $r(s,a)$ and $\mathbf{P}$ the $n \times k \times n$ transition matrix. We use $\mathbf{r_a}$ and $\mathbf{P_a}$ to denote the obvious restrictions of these to action $a$. We define $\mathbf{E}$ to be the $nk \times n$-matrix with a row for each state-action pair and one column per state, with $\mathbf{E}_{sa,t} = P_{sa}(t)$ if $t \neq s$, and $\mathbf{E}_{sa,t} = P_{sa}(t) - 1$ if $t = s$.

Our aim is to find an optimal *policy* that maximizes expected discounted reward. A deterministic policy $\pi : S \rightarrow A$ has *value function* $V^\pi$ satisfying:

$$V^\pi(s) = r(s, \pi(s)) + \gamma \sum_{s'} P_{s\pi(s)}(s')V^\pi(s')$$

or equivalently (slightly abusing subscript $\pi$):

$$\mathbf{V}^\pi = \mathbf{r}_{\mathbf{a}_\pi} + \gamma \mathbf{P}_{\mathbf{a}_\pi}\mathbf{V}^\pi \tag{1}$$

We also define the *Q-function* $Q : S \times A \rightarrow \mathbb{R}$ as:

$$\mathbf{Q}^\pi_\mathbf{a} = \mathbf{r_a} + \gamma \mathbf{P_a}\mathbf{V}^\pi,$$

i.e., the value of executing $\pi$ forward after taking action $a$.

A policy $\pi$ induces a *visitation frequency function* $f^\pi$, where $f^\pi(s,a)$ is the total discounted joint probability of being in state $s$ and taking action $a$. The policy can readily be recovered from $f^\pi$, via $\pi(s,a) = f^\pi(s,a)/\sum_{a'} f^\pi(s,a')$. (For deterministic policies, $\mathbf{f}^\pi_{\mathbf{sa}} = \mathbf{0}$ for all $a$ other than $\pi(s)$.) We use $\mathcal{F}$ to denote the set of valid visitation frequency functions (w.r.t. a fixed MDP), i.e., those satisfying [17]:

$$\gamma \mathbf{E}^\top \mathbf{f} + \alpha = \mathbf{0}. \tag{2}$$

The *optimal value function* $V^*$ satisfies:

$$\alpha \mathbf{V}^* = \mathbf{r}^\top \mathbf{f}^* \tag{3}$$

where $\mathbf{f}^* = \sup_\mathbf{f} \mathbf{r}^\top \mathbf{f}$ [17]. Thus, determining an optimal policy is equivalent to finding optimal frequencies $\mathbf{f}^*$.

### 2.2 Minimax Regret for Imprecise MDPs

A number of researchers have considered the problem of solving imprecisely specified MDPs (see below). Here we focus on the solution of MDPs with imprecise reward functions. Since fully specifying reward functions is difficult, we will often be faced with the problem of computing policies with an incomplete reward specification. Indeed, as we see below, we often explicitly wish to leave parts of a reward function unelicited (or otherwise unassessed). Formally we assume that $\mathbf{r} \in \mathcal{R}$, where the feasible reward set $\mathcal{R}$ reflects current knowledge of the reward. These could reflect: prior bounds specified by a user or domain expert; constraints that emerge from an elicitation process (as discussed below); or constraints that arise from observations of user behavior (as in inverse RL [15]). In all of these situations, we are unlikely to have full reward information. Thus we require a criterion by which to compare policies in an imprecise-reward MDP.

We adopt the *minimax regret criterion*, originally suggested (though not endorsed) by Savage [18], and applied with some success in non-sequential decision problems [6, 7]. Let $\mathcal{R}$ be the set of feasible reward functions. Minimax regret can be defined in three stages:

$$R(\mathbf{f}, \mathbf{r}) = \max_{\mathbf{g} \in \mathcal{F}} \quad \mathbf{r} \cdot \mathbf{g} - \mathbf{r} \cdot \mathbf{f} \tag{4}$$

$$MR(\mathbf{f}, \mathcal{R}) = \max_{\mathbf{r} \in \mathcal{R}} \quad R(\mathbf{f}, \mathbf{r}) \tag{5}$$

$$MMR(\mathcal{R}) = \min_{\mathbf{f} \in \mathcal{F}} \quad MR(\mathbf{f}, \mathcal{R}) \tag{6}$$

$R(\mathbf{f}, \mathbf{r})$ is the *regret* of policy $\mathbf{f}$ (as represented by its visitation frequencies) relative to reward function $\mathbf{r}$: it is simply the loss or difference in value between $\mathbf{f}$ and the optimal policy under $\mathbf{r}$. $MR(\mathbf{f}, \mathcal{R})$ is the *maximum regret* of $\mathbf{f}$ w.r.t. feasible reward set $\mathcal{R}$. Should we chose a policy with visitation frequencies $\mathbf{f}$, $MR(\mathbf{f}, \mathcal{R})$ represents the worst-case loss over all possible realizations of the reward function; i.e., the regret incurred in the presence of an *adversary* who chooses the $\mathbf{r}$ from $\mathcal{R}$ to maximize our loss. Finally, in the presence of such an adversary, we wish to minimize this max regret: $MMR(\mathcal{R})$ is the *minimax regret* of feasible reward set $\mathcal{R}$. This can be viewed as a game between a decision maker choosing $\mathbf{f}$ who wants to minimize loss relative to the optimal policy, and an adversary who chooses a reward to maximize this loss given the decision maker's choice of policy. Any $\mathbf{f}^*$ that minimizes max regret is a *minimax optimal policy*, while the $\mathbf{r}$ that maximizes its regret is the *witness* or *adversarial reward function*, and the optimal policy $\mathbf{g}$ for $\mathbf{r}$ is the *witness* or *adversarial policy*.

Minimax regret has a variety of desirable properties relative to other robust decision criteria [6]. Compared to Bayesian methods that compute expected value using a prior over $\mathcal{R}$ [3, 8], minimax regret provides worst-case bounds on loss. Specifically, let $\mathbf{f}$ be the minimax regret optimal visitation frequencies and let $\delta$ be the max regret achieved by $\mathbf{f}$; then, given any instantiation of $\mathbf{r}$, no policy will outperform $\mathbf{f}$ by more than $\delta$ w.r.t. expected value. Minimax optimal decisions can often be computed more effectively than decisions that maximize expected value w.r.t. to some prior. Finally, it has been shown to be a very effective criterion for driving elicitation in one-shot problems [6, 7].

### 2.3 Robust Optimization for Imprecise MDPs

Most work on robust optimization for imprecisely specified MDPs adopts the *maximin criterion*, producing policies with maximum *security level* or worst-case value [1, 13, 14, 16]. Restricting attention to imprecise rewards, the maximin value is given by:

$$MMN(\mathcal{R}) = \max_{\mathbf{f}\in\mathcal{F}} \min_{\mathbf{r}\in\mathcal{R}} \quad \mathbf{r}\cdot\mathbf{f} \tag{7}$$

Most models are defined for uncertainty in any MDP parameters, but algorithmic work has focused on uncertainty in the transition function, and the of eliciting information about transition functions or rewards is left unaddressed. Robust policies can be computed for uncertain transition functions using the maximin criterion by decomposing the problem across time-steps and using dynamic programming and an efficient suboptimization to find the worst case transition function [1, 13, 16]. McMahan, Gordon, and Blum [14] develop a linear programming approach to efficiently compute the maximin value of an MDP (we empirically compare this approach to ours below). Delage and Mannor [9] address the problem of uncertainty over reward functions (and transition functions) in the presence of prior information, using a percentile criterion, which can be somewhat less pessimistic than maximin. They also contribute a method for eliciting rewards using sampling to approximate the expected value of information of noisy information about a point in reward space. The percentile approach is neither fully Bayesian nor does it offer a bound on performance. Zhang and Parkes ([20]) also adopt maximin in a model that assumes an inverse reinforcement learning setting for *policy teaching*. The approach is essentially a form of reward elicitation which the queries are changes to a student's reward, and information is gained by observing change in the student's behavior.

Generally, the *maximin* criterion leads to conservative policies by optimizing against the worst possible instantiation of $\mathbf{r}$ (as we will see below). Minimax regret offers a more intuitive measure of performance by assessing the policy *ex post* and making comparisons only w.r.t. specific reward realizations. Thus, policy $\pi$ is penalized on reward $\mathbf{r}$ only if there exists a $\pi'$ that has higher value w.r.t. $\mathbf{r}$ itself.

## 3 Minimax Regret Computation

As discussed above, maximin is amenable to dynamic programming since it can be decomposed over decision stages. This decomposition does not appear tenable for minimax regret since it grants the adversary too much power by allowing rewards to be set independently at each stage (though see our discussion of future work below). Following the formulations for non-sequential problems developed in [6, 7], we instead formulate the optimization using a series of linear (LPs) and mixed integer programs (MIPs) that enforce a consistent choice of reward across time.

Assume feasible reward set $\mathcal{R}$ is represented by a convex polytope $\mathbf{Cr} \le \mathbf{d}$, which we assume to be bounded. The constraints on $\mathbf{r}$ arise as discussed above (prior bounds, elicitation, or behavioral observation). Minimax regret can then be expressed as following minimax program:

$$\begin{aligned} \min_{\mathbf{f}} \quad & \max_{\mathbf{g}} \max_{\mathbf{r}} \quad \mathbf{r}\cdot\mathbf{g} - \mathbf{r}\cdot\mathbf{f} \\ \text{subject to:} \quad & \gamma\mathbf{E}^\top\mathbf{f} + \alpha = \mathbf{0} \\ & \gamma\mathbf{E}^\top\mathbf{g} + \alpha = \mathbf{0} \\ & \mathbf{Cr} \le \mathbf{d} \end{aligned}$$

This is equivalent to a minimization:

$$\begin{aligned} \underset{\mathbf{f},\delta}{\text{minimize}} \quad & \delta \\ \text{subject to:} \quad & \mathbf{r}\cdot\mathbf{g} - \mathbf{r}\cdot\mathbf{f} \le \delta \quad \forall\, \mathbf{g}\in\mathcal{F}, \mathbf{r}\in\mathcal{R} \\ & \gamma\mathbf{E}^\top\mathbf{f} + \alpha = \mathbf{0} \end{aligned} \tag{8}$$

This corresponds to the standard dual LP formulation of an MDP with the addition of adversarial policy constraints. The infinite number of constraints can be reduced: first we need only retain as potentially active those constraints for vertices of polytope $\mathcal{R}$; and for any $\mathbf{r}\in\mathcal{R}$, we only require the constraint corresponding to its optimal policy $\mathbf{g}^*_{\mathbf{r}}$. However, vertex enumeration is not feasible; so we apply Benders' decomposition [2] to iteratively generate constraints.

At each iteration, two optimizations are solved. The *master problem* solves a relaxation of program (8) using only a small subset of the constraints, corresponding to a subset *Gen* of all $\langle \mathbf{g},\mathbf{r}\rangle$ pairs; we call these *generated constraints*. Initially, this set is arbitrary (e.g., empty). Intuitively, in the game against the adversary, this restricts the adversary to choosing witnesses (i.e., $\langle \mathbf{g},\mathbf{r}\rangle$ pairs) from *Gen*.

Let $\mathbf{f}$ be the solution to the current master problem and $MMR'(\mathcal{R})$ its objective value (i.e., minimax regret in the presence of the restricted adversary). The *subproblem* generates the maximally violated constraint relative to $\mathbf{f}$. In other words, we compute $MR(\mathbf{f},\mathcal{R})$; its solution determines the witness points $\langle \mathbf{g},\mathbf{r}\rangle$ by removing restrictions on the adversary. If $MR(\mathbf{f}, \mathcal{R}) = MMR'(\mathcal{R})$ then the constraint for $\langle \mathbf{g}, \mathbf{r} \rangle$ is satisfied at the current solution, and indeed all unexpressed constraints must be satisfied as well. The process then terminates with minimax optimal solution $\mathbf{f}$. Otherwise, $MR(\mathbf{f}, \mathcal{R}) > MMR'(\mathcal{R})$, implying that the constraint for $\langle \mathbf{g}, \mathbf{r} \rangle$ is violated in the current relaxation (indeed, it is the maximally violated such constraint). So it is added to *Gen* and the process repeats.

Computation of $MR(\mathbf{f}, \mathcal{R})$ is realized by the following MIP, using value and Q-functions:[1]

$$\begin{aligned}
\underset{\mathbf{Q},\mathbf{V},\mathbf{I},\mathbf{r}}{\text{maximize}} \quad & \alpha \cdot \mathbf{V} - \mathbf{r} \cdot \mathbf{f} && && (9)\\
\text{subject to:} \quad & \mathbf{Q_a} = \mathbf{r_a} + \gamma \mathbf{P_a} \mathbf{V} && \forall\, a \in A &&\\
& \mathbf{V} \geq \mathbf{Q_a} && \forall\, a \in A && (10)\\
& \mathbf{V} \leq (\mathbf{1} - \mathbf{I_a})\mathbf{M_a} + \mathbf{Q_a} && \forall\, a \in A && (11)\\
& \mathbf{Cr} \leq \mathbf{d} && &&\\
& \sum_a \mathbf{I_a} = \mathbf{1} && && (12)\\
& \mathbf{I_a}(\mathbf{s}) \in \{\mathbf{0}, \mathbf{1}\} && \forall a, s && (13)\\
& \mathbf{M_a} = \mathbf{M}^{\top} - \mathbf{M}^{\perp}_{\mathbf{a}} && &&
\end{aligned}$$

Here $\mathbf{I}$ represents the adversary's policy, with $\mathbf{I_a}(\mathbf{s})$ denoting the probability of action $a$ being taken at state $s$ (constraints (12) and (13) restrict it to be deterministic). Constraints (10) and (11) ensure that the optimal value $V(s) = Q(s, a)$ for a single action $a$. We ensure a tight $\mathbf{M}^{\perp}_{\mathbf{a}}$ by setting $\mathbf{M}^{\top}$ to be the optimal value function $\mathbf{V}^{\top}$ of the optimal policy with respect to the best setting of each individual reward point and $\mathbf{M}^{\perp}_{\mathbf{a}}$ to be the Q-value $\mathbf{Q}^{\perp}_{\mathbf{a}}$ of the optimal policy with respect to the worst point-wise setting of rewards (the resulting rewards need not be feasible).

The subproblem does not directly produce a witness pair $\langle \mathbf{g}_i, \mathbf{r}_i \rangle$ for the master constraint set; instead it provides $\mathbf{r}_i$ and $\mathbf{V}_i$. However, we do not need access to $\mathbf{g}_i$ directly; the constraint can be posted using the reward function $\mathbf{r}_i$ and the value $\alpha \cdot \mathbf{V}_i$, since $\alpha \cdot \mathbf{V}_i = \mathbf{r}_i \cdot \mathbf{g}_i$ (and $\mathbf{g}_i$ is required to determine this adversarial value in the posted constraint).

In practice we have found that the iterative constraint generation converges quickly, with relatively few constraints required to determine minimax regret (see Sec. 5). However, the computational cost per iteration can be quite high. This is due exclusively to the subproblem optimization, which requires the solution of a MIP with a large number of integer variables, one per state-action pair. The master problem optimization, by contrast, is extremely effective (since it is basically a standard MDP linear program). This suggests examination of approximations to the subproblem, i.e., the computation of max regret $MR(\mathbf{f}, \mathcal{R})$. This is also motivated by our focus on reward elicitation. We wish to use minimax regret to drive query selection: our aim is not to compute minimax regret for its own sake, but to determine which state-action pairs should be queried, i.e., which have the potential to reduce minimax regret. The visitation frequencies used by our heuristics need not correspond to exact minimax optimal policy.

We have explored several promising alternatives, including an alternating optimization model that computes an adversarial policy (for a fixed reward) and an adversarial reward (for a fixed policy). This reduces the quadratic optimization for max regret to a sequence of LPs. An simpler approximation is explored here (which performs as well in practice): we solve the LP relaxation of the MIP by removing the integrality constraints (13) on the binary policy indicators. The value function $\mathbf{V}$ resulting from this relaxation does not accurately reflect the (now stochastic) adversarial policy: $\mathbf{V}$ may include a fraction of the big-M term due to constraint (10). However, the reward function $\mathbf{r}$ selected remains in the feasible set, and, empirically, the optimal value function for $\mathbf{r}$ yields a solution to the subproblem that is close to optimal.[2] Since the reward is valid choice, this solution is guaranteed to be a lower bound on the solution to the subproblem. When this approximate subproblem solution is used in constraint generation, convergence is no longer guaranteed; however, the solution to the master problem represents a valid lower bound on minimax regret.

## 4 Reward Elicitation

Reward elicitation and assessment can proceed in a variety of ways. Many different query forms can be adopted for user interaction. Similarly, observed user behavior can be used to induce constraints on the reward function under assumptions of user "optimality" [15]. In this work, we focus on simple *bound queries*, though our strategies can be adapted to more general query types. We discuss some of these below.[3]

We assume that $\mathcal{R}$ is given by upper and lower bounds on $r(s, a)$ for each state-action pair. A bound query takes the form *"Is* $r(s, a) \geq b$*?"* where $b$ lies between the upper and lower bound on $r(s, a)$. While this appears to require a direct, quantitative assessment of value/reward by the user, it can be recast as a *standard gamble* [10], a device used in decision analysis to reduce this to preference query over two outcomes (one of which is stochastic). For simplicity, we express it in this bound form. Unlike reward queries [9], which require a direct assessment of $r(s, a)$, bound queries require only a yes-no response and are less cognitively demanding. A response tightens either the upper or lower

[1] Specifying max regret in terms of visitation frequencies (i.e., the standard dual MDP formulation) gives rise to a non-convex quadratic program. Regret maximization does not lend itself to a natural, linear primal formulation.

[2] Finding the optimal value function for $\mathbf{r}$ requires solving a standard MDP LP.

[3] We allow reward queries about any state-action pair, in contrast to online RL formalisms, in which information can be gleaned only about the reward (and dynamics) at the current state. As such, we face no exploration-exploitation tradeoff.

bound on $r(s,a)$.[4]

Bound queries offer a natural starting point for the investigation of reward elicitation. Of course, many alternative query modes can be used, with the sequential nature of the MDP setting opening up choices that don't exist in one-shot settings. These include the direct comparison of policies; comparison of (full or partial) state-action trajectories or distributions over trajectories; and comparisons of outcomes in factored reward models. Trajectory comparisons can be facilitated by using counts of relevant (or reward-bearing) events as dictated by a factored reward model for example. These query forms should prove useful and even more cognitively penetrable. However, the principles and heuristics espoused below can be adapted to these settings.

There are many ways to select the point $(s,a)$ at which to ask a bound query. We explore some simple myopic heuristic criteria that are very easy to compute, are based on criteria suggested in [6]. The first selection heuristic is called *halve largest gap (HLG)*, which selects the point $(s,a)$ with the largest gap between its upper and lower bound. Formally, we define the gap $\Delta(s,a)$ and largest gap by:

$$\Delta(s,a) = \max_{r' \in \mathcal{R}} r'(s,a) - \min_{r \in \mathcal{R}} r(s,a)$$

$$\operatorname*{argmax}_{a^* \in A, s^* \in S} \Delta(s^*, a^*)$$

The second selection heuristic is the *current solution (CS) strategy*, and uses the visitation frequencies from the minimax optimal solution $\mathbf{f}$ or the adversarial witness $\mathbf{g}$ to weight each gap. Intuitively, if a query involves a reward parameter that influences the value of neither $\mathbf{f}$ nor $\mathbf{g}$, minimax regret will not be reduced, and visitation frequencies quantify the *degree* of influence. Formally CS selects the point:

$$\operatorname*{argmax}_{a^* \in A, s^* \in S} \max\{f(s^*,a^*)\Delta(s^*,a^*),\ g(s^*,a^*)\Delta(s^*,a^*)\}.$$

Given the selected $(s^*, a^*)$, bound $b$ in the query is set to the midpoint of the interval for $r(s^*, a^*)$. Thus either response will reduce the interval by half. It is easy to apply CS to the maximin criterion as well, using the visitation frequencies associated with the maximin policy.

# 5 Experiments

We assess the general performance of our approach using a set of randomly generated MDPs and specific MDPs arising in an *autonomic computing* setting. We assess scalability of our procedures, as well as the effectiveness of minimax regret as a driver of elicitation.

We first consider randomly generated MDPs. We impose structure on the MDP by creating a semi-sparse transition function: for each $(s,a)$-pair, $\lceil \log n \rceil$ reachable states are drawn uniformly and a Gaussian is used to generate transition probabilities. We use a uniform initial state distribution $\alpha$ and discount factor $\gamma = 0.95$. The true reward is drawn uniformly from a fixed interval and uncertainty w.r.t. this true (but unknown) reward is created by bounding each $(s,a)$-pair independently with bounds drawn randomly: thus the set of feasible rewards forms a hyper-rectangle.[5]

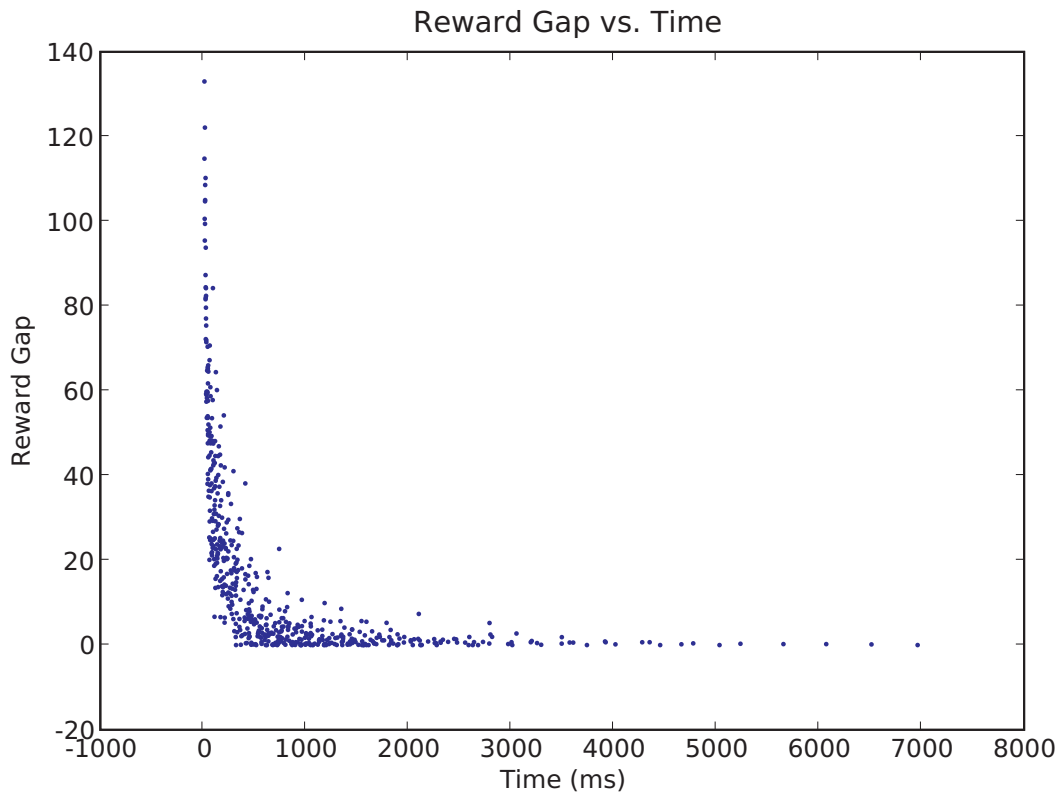

Figure 1: Reduction in regret gap during constraint generation.

## 5.1 Computational Efficiency

To measure the performance of minimax regret computation, we first examine the constraint generation procedure. Fig. 1 plots the *regret gap* between the master problem value and subproblem value at each iteration versus the time (in ms.) to reach that iteration. Results are shown for 20 randomly generated MDPs with ten states and five actions. Fig. 2 shows how minimax regret computation time increases with the size of the MDP (5 actions, varying number of states). Constraint generation using the MIP formulation scales super-linearly, hence computing minimax regret exactly is only feasible for small MDPs using this formulation; by comparison the linear relaxation is far more efficient.[6] On the other hand, minimax regret computation has very favorable anytime behavior, as exhibited in Fig. 1. During constraint generation, the regret gap shrinks very quickly early on. If exact minimax regret is not needed, this property allows for fast approximation.

## 5.2 Approximation Error

To evaluate the linear relaxation scheme for max regret, we generated random MDPs, varying the number of states. Fig. 3 shows average relative error over 100 runs. The approximation performs well and, encouragingly, error does not increase with the size of the MDP. We also evaluate its impact on minimax regret when used to generate violated constraints. Fig. 3 also shows relative error for minimax regret to be small, well under 10% on average.

[4] Indifference (e.g., "I'm not sure") can also be handled by constraining bounds to be within $\varepsilon$ of the query point.

[5] CPLEX 11 is used for all MIPS and LPs, and all code run on a PowerEdge 2950 server with dual quad-core Intel E5355 CPUs.

[6] Of note, the computations shown here are using the initial reward uncertainty. As queries refine the reward polytope, regret computation becomes faster in general. This has positive impli-

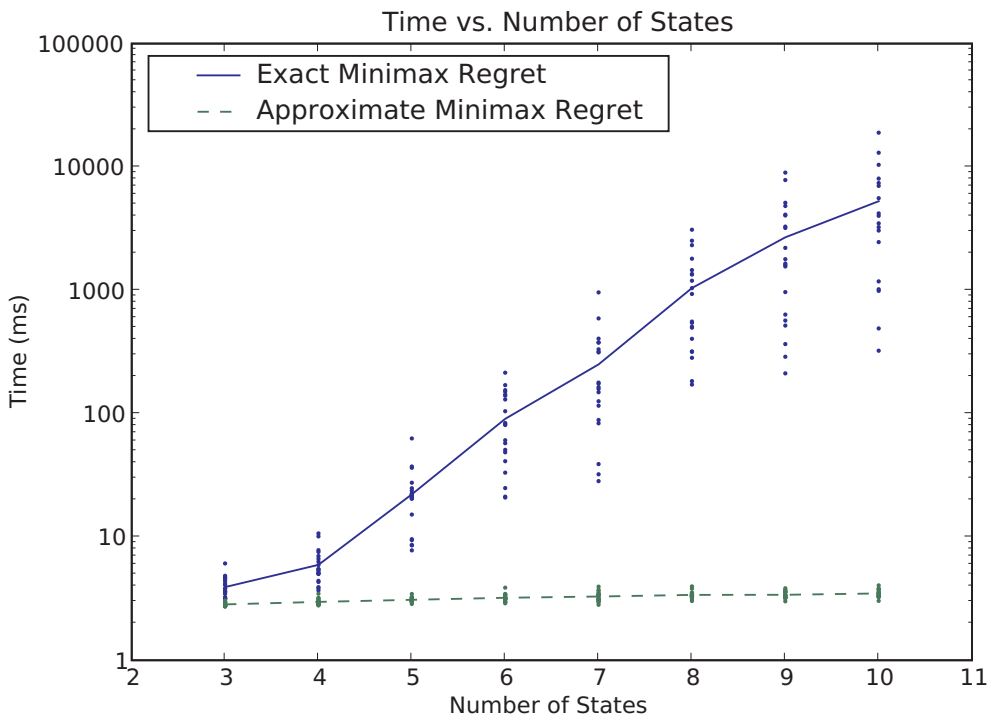

Figure 2: Scaling of constraint generation with number of states.

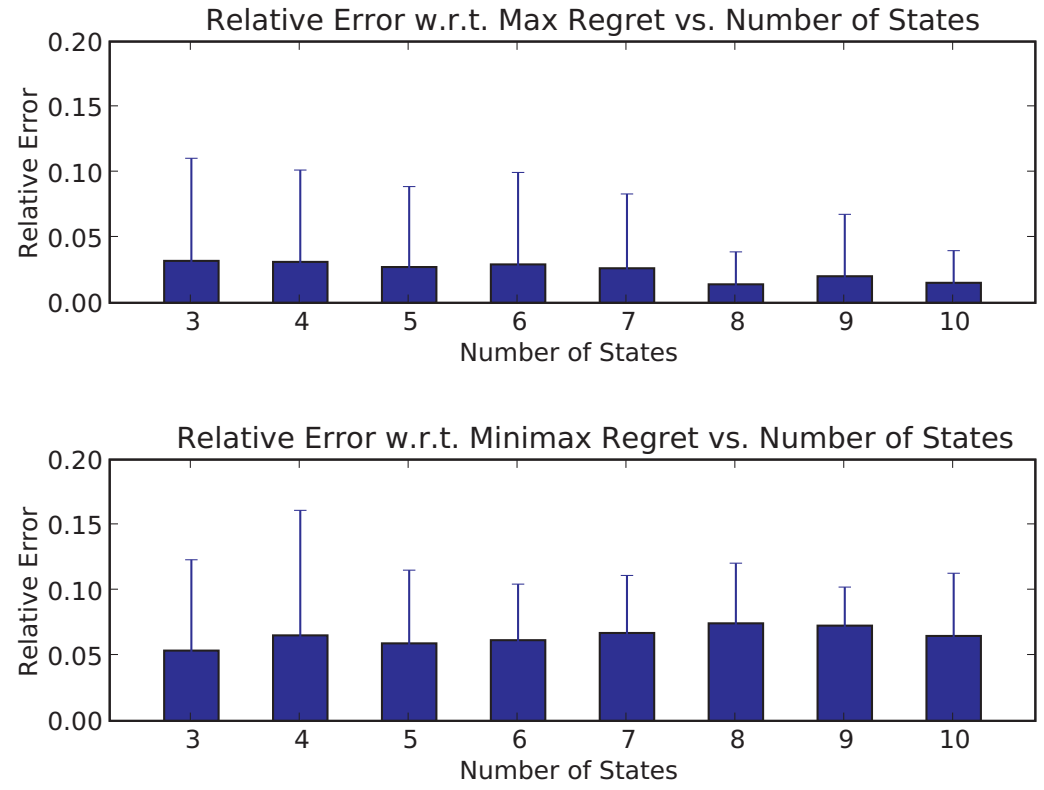

Figure 3: Relative approximation error of linear relaxation

### 5.3 Elicitation Effectiveness

We analyzed the effectiveness of our regret-based elicitation procedure by comparing it with the maximin criterion. We implemented a variation of the Double Oracle maximin algorithm developed by McMahan, Gordon & Blum [14]. The computation time for maximin is significantly less the that of minimax regret—this is expected since maximin requires only the solution of a pair of linear programs.

We use both maximin and minimax regret to compute policies at each step of preference elicitation, and paired each with the current solution (CS) and halve largest gap (HLG) query strategies, giving four elicitation procedures: MMR-HLG (policies are computed using regret, queries generated by HLG); MMR-CS (regret policies, CS queries); MM-HLG (maximin policies, HLG queries); and MM-CS (maximin policies, CS queries). We assess each procedure by measuring the quality of the policies produced after each query, using the following metrics: (a) its maximin value given the current (remaining) reward uncertainty; (b) its max regret given the current (remaining) reward uncertainty; and (c) its true regret (i.e., loss w.r.t. the optimal policy for the true reward function $\mathbf{r}$, where $\mathbf{r}$ is used to generate query responses). Minimax regret is the most critical since it provides the strongest guarantees; but we compare to maximin value as well, since maximin policies are optimizing against a very different robustness measure. True cations for anytime computation.

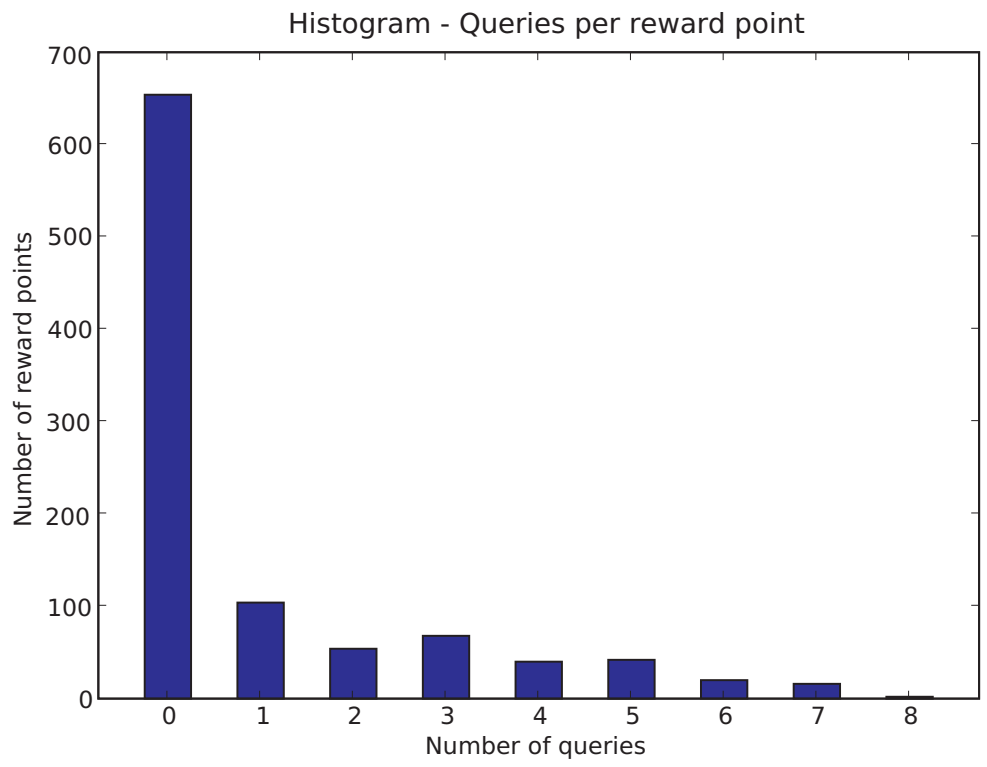

Figure 5: Number of queries at each state-action pair using MMR-CS.

regret is not available in practice; but it gives an indication of how good the resulting policies actually are (as opposed to a worst-case bound).

Fig. 4 show the results of the comparison on each measure. MMR-CS performs extremely well on all measures. Somewhat surprisingly, it outperforms MM-CS and MM-HLG w.r.t. maximin value (except at the very early stages). Even though the maximin procedures are optimizing maximin value, MMR-CS asks much more informative queries, allowing for a larger reduction in reward uncertainty at the most relevant state-action pairs. This ability of MMR-CS to identify the highest impact reward points becomes clearer still when we examine how much reduction there is in reward intervals over the course of elicitation. Let $\chi$ measure the sum of the length of the reward intervals. At the end of elicitation, MMR-HLG reduces $\chi$ to 15.6% of its original value (averaged over the 20 MDPs), while MMR-CS only reduces $\chi$ to 67.8 % of its original value. MMR-CS is effectively eliminating regret while leaving a large amount of uncertainty. Fig. 5 illustrates this using a histogram of the number of queries asked by MMR-CS about each of the 1000 possible state-action pairs.[7] We see that MMR-CS asks no queries about the majority of state-action pairs, and asks quite a few queries (up to eight) about a small number of "high impact" pairs.

Fig. 4(b) shows that MMR-CS is able to reduce regret to zero (i.e., find an optimal policy) after less than 100 queries on average. Recall that the MDP has 50 reward parameters (state-action pairs), so on average, less than two queries per parameter are required to find a provably optimal policy. The minimax regret policies also outperform the maximin policies by a wide margin with respect to true regret (Fig. 4(c)). With the CS heuristic, a near-optimal policy is found after fewer than 50 queries (less than one query per parameter), though to *prove* that the policy is near-optimal requires further queries (to reduce minimax regret).

It is worth noting that during preference elicitation, HLG does not require that minimax regret actually be computed.

[7] 20 MDPs with 10 states, 5 actions each.

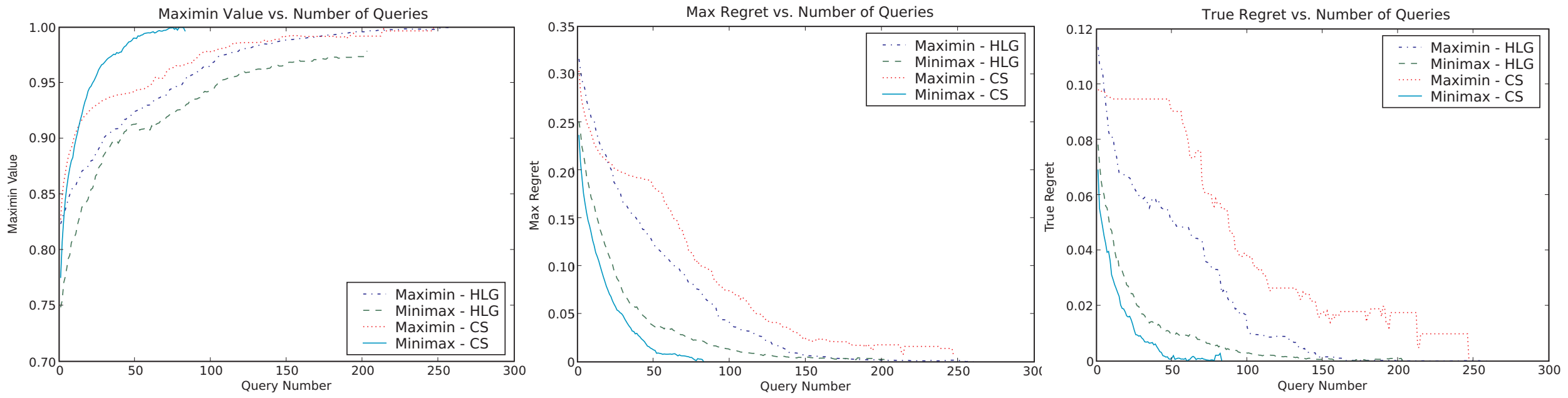

(a) Maximin (b) Max Regret (c) True Regret

Figure 4: Reward elicitation with randomly generated MDPs.

Minimax regret is only necessary to assess when to stop the elicitation process (i.e., to determine if minimax regret has dropped to an acceptable level). One possible modification to reduce the time between queries is to only compute minimax regret after every $k$ queries. Of course, the HLG strategy will lead to a slower reduction in true regret and minimax regret as shown in Figs. 4(b) and 4(c).

To further evaluate our approach we elicit the reward function for an *autonomic computing scenario* [4] in which we must allocate computing or storage resources to application servers as their client demands change over time. We assume $k$ *application server elements* and $N$ *units of resource* available to be assigned to the servers (plus a "zero resource"). An allocation $\mathbf{n} = \langle n_1 \ldots n_k \rangle$ must satisfy $\sum_i^k n_i < N$. There are $D$ *demand levels* at which each server can operate, reflecting client demands. A *demand state* $\mathbf{d} = \langle d_1 \ldots d_k \rangle$ specifies the current demand for each server. A state of the MDP comprises the current resource allocation and the current demand state: $s = \langle \mathbf{n}, \mathbf{d} \rangle$. Actions are new allocations $\mathbf{m} = \langle m_1 \ldots m_k \rangle$ of the $N$ resources to the $k$ servers. Reward $r(\mathbf{n}, \mathbf{d}, \mathbf{m}) = u(\mathbf{n}, \mathbf{d}) - c(\mathbf{n}, \mathbf{d}, \mathbf{m})$ decomposes as follows. Utility $u(\mathbf{n}, \mathbf{d})$ is the sum of server utilities $u_i(n_i, d_i)$. The MDP is initially specified with strict uncertainty over the utilities $u_i$ however, we assume that each utility function $u_i$ is monotonic non-decreasing in demand and resource level. The cost $c(\mathbf{n}, \mathbf{d}, \mathbf{m})$ is the sum of the costs of taking away one unit of resource from each server at any stage. Uncertainty in demand is exogenous and the action in the current state uniquely determines the allocation in the next state. Thus the transition function is composed of $k$ Markov chains $\Pr(d'_i \mid d_i), i \leq k$. Reward specification in this context is inherently distributed and quite difficult: the local utility function $u_i$ for server $i$ has no convenient closed form. Server $i$ can respond only to queries about the utility it gets from a *specific* resource allocation level, and this requires intensive optimization and simulation on the part of the server [4]; hence minimizing the number of such queries is critical.

We constructed a small instance of the autonomic computing scenario with 2 servers, 3 demand levels and 3 (indivisible) units of resource. The combined state space of both servers includes $3^2$ demand levels and 10 possible allocations of resources leading to 90 states and 10 actions. We modeled the uncertainty over rewards using a hyper-rectangle as with the random MDPs. We compared elicitation approaches as above, this time using the linear relaxation to compute minimax regret (each minimax computation takes under 3s.). Fig. 6 shows that MMR-CS again outperforms the maximin criterion on each measure. Minimax regret and true regret fall to almost zero after 200 queries. Recall that the autonomic MDP had 900 state-action pairs—the additional problem structure results in fewer than 0.25 queries being asked for each state-action pair. In fact, on average MMR-CS only asks about 106.5 distinct state-action pairs, only examining 12% of the reward space. By comparison, the queries chosen by the MM-CS strategy cover just over 68% of the reward space. As with random MDPs, minimax regret quickly reduces regret because it focuses queries on the "high impact" state-action pairs.

Overall, our regret-based approach is quite appealing from the perspective of reward elicitation. While the regret computation is more computationally intensive than other criteria, it provides arguably much more natural decisions in the face of reward uncertainty. More importantly, from the perspective of elicitation, it is much more attractive than maximin w.r.t. the number of queries required to produce high-quality policies. As long as interaction time (time between queries) remains reasonable, reducing user burden (or other computational costs required to answer queries) is our primary goal.

## 6 Conclusions & Future Work

We have developed an approach to reward elicitation in MDPs that eases the burden of reward function elicitation. Minimax regret not only offers robust policies in the face of reward uncertainty, but we've shown it also allows one to focus elicitation attention on the most important aspects of the reward function. While the computational costs are significant, it is an extremely effective driver of elicitation, thus reducing the (more important) cognitive or computational cost of reward determination. Furthermore, it lends itself to anytime approximation.

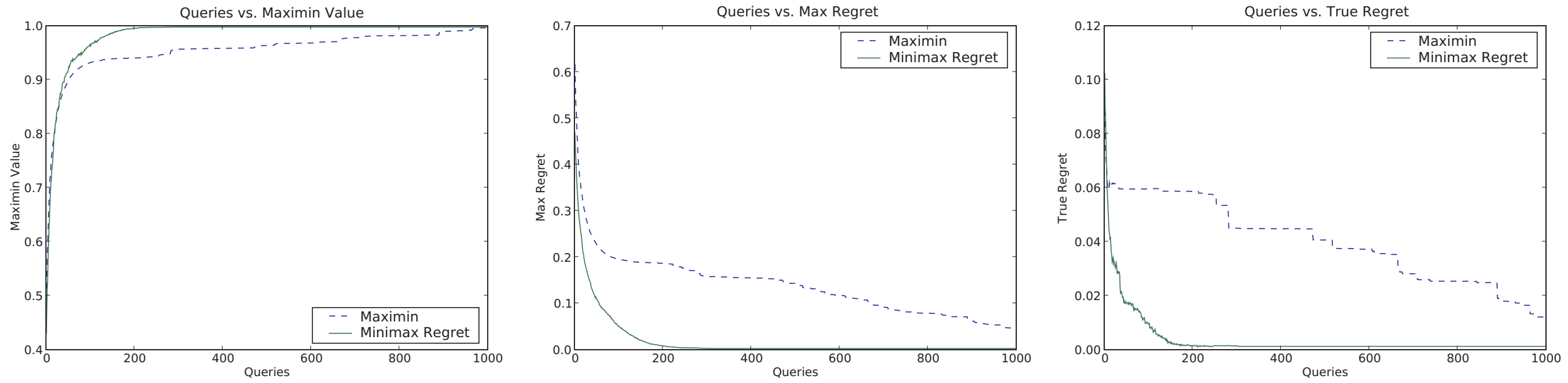

(a) Maximin (b) Max Regret (c) True Regret

Figure 6: Elicitation of reward in autonomic computing domain

The somewhat preliminary nature of this work leaves many interesting directions for future research. Perhaps most interesting is the development of more informative and intuitive queries that capture the sequential nature of the elicitation problem. Direct comparison of policies allows one to distinguish value from reward, but are cognitively demanding. Trajectory comparison similar distinguishes value, but may contain irrelevant detail. However, trajectory summaries (e.g., counts of relevant reward bearing events) may be more perspicuous, and could be generated to reflect expected "event counts" given a policy. Other forms of queries should also prove valuable, but all exploit the basic idea embodied by minimax regret and the current solution heuristic. Another direction for improving elicitation is to incorporate *implicit* information in a manner similar to policy teaching [20]. Inverse RL [15] can be also used to translate observed behavior into constraints on reward. Some Bayesian models [6, 8] allow noisy query responses and adding this to our regret model is another important direction. Two approaches include: approximate indifference constraints and regret-based sensitivity analysis. The efficiency of the minimax regret computation remains an important research topic. We are exploring the use of dynamic programming to generate linear representations of the best policies over all regions of reward space (much like POMDPs) which can greatly assist max regret computation. We are also exploring techniques that exploit factored MDP structure using LP approaches [12].